\documentclass[conference]{IEEEtran}
\IEEEoverridecommandlockouts
\usepackage{cite}
\usepackage{amsmath,amssymb,amsfonts}
\usepackage{algorithmic}
\usepackage{graphicx}
\usepackage{textcomp}
\usepackage{xcolor}
\def\BibTeX{{\rm B\kern-.05em{\sc i\kern-.025em b}\kern-.08em
    T\kern-.1667em\lower.7ex\hbox{E}\kern-.125emX}}
\begin{document}
	
	\bibliographystyle{IEEEtran}
	
\title{From Data Quality to Model Quality: 
\\ an Exploratory Study on Deep Learning}

\author{\IEEEauthorblockN{Tianxing He$^1$, Shengcheng Yu$^1$, Ziyuan Wang$^{2*}$, Jieqiong Li$^1$, Zhenyu Chen$^{1*}$}
	\IEEEauthorblockA{$^1$State Key Lab of Novel Software Technology, Nanjing University \\
	$^2$School of Computer Science, Nanjing University of Posts and Telecommunications \\
		 $^*$corresponding authors: wangziyuan@njupt.edu.cn, zychen@nju.edu.cn}}
		 
\maketitle

\begin{abstract}
Nowadays, people strive to improve the accuracy of deep learning models. However, very little work has focused on the quality of data sets. In fact, data quality determines model quality. Therefore, it is important for us to make research on how data quality affects on model quality. In this paper, we mainly consider four aspects of data quality, including Dataset Equilibrium, Dataset Size, Quality of Label, Dataset Contamination. We deign experiment on MNIST and Cifar-10 and try to find out the influence the four aspects make on model quality. Experimental results show that four aspects all have decisive impact on the quality of models. It means that decrease in data quality in these aspects will reduce the accuracy of model.
\end{abstract}

\section{Introduction}
 For a long time, data and training model play essential roles in the field of deep learning \cite{shi2019deepgini}. People have devoted much to the study of training model while they neglected the study of the training set itself. It is well known that the quality of datasets plays a decisive role in the final training results. For example, It is impossible to train a high-quality training model with very high test accuracy on a dataset with a random set. So, what are the aspects that measure the quality of a dataset \cite{Gu2019DQ} and how does them influence the quality of data set? Solving these problems is not only helpful for people to tell the quality of datasets, but also for people to balance training costs and accuracy in developing depth learning applications. Moreover, it may make contributions to the research of Adversarial examples and Data Augmentation. This paper firstly focuses on the quality of deep learning image dataset, which will have a positive impact on future research work.

In this paper, we mainly consider four aspects of data quality, including：Dataset Equilibrium, Dataset Size, Quality of Label, Dataset Contamination. We design experiments to find out the influence of each aspect on the quality of data sets and the influence curve of added errors or noises on the accuracy of test sets. We use MNIST \cite{lecun1998mnist} \cite{platt1999using} to train Lenet-5 and use Cifar-10 \cite{krizhevsky2009learning} to train Resnet20 \cite{he2016deep} and NetworkInNetwork \cite{lin2013network} in these experiments.

We obtained a series of discoveries from experiments:
    \begin{itemize}
    \item Label errors are the most harmful aspect to datasets. A dataset with more than 20\% label errors cannot be used for training.
    
    \item Some noises can help raise the accuracy of the testing set, while the change of brightness seems helpful.
    
    \item When the number of pictures in the image dataset reaches a certain amount, the improvement of the accuracy of the testing set is relatively limited. Before that, the effect was remarkable.

	\item When we randomly delete a class from the training set or alter the label of one class to that of another class, the reduction of the accuracy of the test set is roughly equal. However, when we delete all pictures of a class, its impact on the model varies with the content of that class. 
    \end{itemize}
    
By doing these, we can easily answer what are the aspects that measure the quality of a dataset and how does them influence the quality of data set. It makes contributions to data augmentation and adversarial example researches and can help people balance costs and accuracy.

\section{Data Quality Aspects}
In this paper, we mainly consider four aspects of data quality which based on Artificial intelligence - Assessment specification for deep learning algorithms.

\textbf{Dataset Equilibrium:} Dataset equilibrium refers to equilibrium degree of samples among classes and deviation of the sample distribution. E.g: We delete all data of one specific category. For example, we delete all ``1''s or ``2''s in the training set of MNIST to see the effect of the model when identifying the deleted digit and  the undeleted digits.

\textbf{Dataset Size:} Dataset size is usually measured by the number of samples, large-scale datasets usually have better sample diversity. E.g.: We modify the dataset size by randomly delete specific percent of data in the training set.

\textbf{Quality of Label:} Quality of label refers to whether the labels of the dataset are complete and accurate. E.g: We randomly change the label to a wrong one, and try different ratio of changed labels, to see the effect on the model robustness.

\textbf{Dataset Contamination:} Dataset contamination refers to the degree of malicious data artificially added to index datasets. E.g: We use different methods including modify contrast, adding noise, etc., to add some contamination to the images, to see the effect on the model robustness.

\section{Experiment}

We conduct our study in order to solve the following research questions:
\begin{itemize}
    \item \textbf{RQ1:} Whether training dataset equilibrium will affect the accuracy of the model.
    \item \textbf{RQ2:} Whether training dataset size will affect the accuracy of the model.
    \item \textbf{RQ3:} Whether the quality of label in the training dataset will affect the accuracy of the model.
    \item \textbf{RQ4:} Whether training dataset contamination will affect the accuracy of the model.
\end{itemize}

\subsection{Experiment setup}
Google researchers have proposed that when the depth of the model is enough, the capacity of neural networks is sufficient for memorizing the entire dataset \cite{zhang2016understanding}, so we will train the model until the accuracy of the training set reaches 100\%. We use lenet-5 for MNIST , Resnet-20 and NetworkInNetwork for Cifar-10.

\textit{Experiment setup for RQ1:} To evaluate the influence of dataset equilibrium to the quality of model, For MNIST, we delete specific classification one by one. For example, we delete all “0” class from the training set. And for Cifar-10, We delete specific classification one by one Or change the labels of one classification to that of another.

\textit{Experiment setup for RQ2:} To evaluate the influence of dataset size to the quality of model, For MNIST, We randomly delete a specific percentage of the images in the training set. (10\% to 100\% at intervals of 10\% and 1\% to 20\% at intervals of 1\%). And for Cifar-10, We delete a specific percentage of the images in the training set. (20\% to 80\% at intervals of 10\% and 85\% to 95\% at intervals of 5\%).

\textit{Experiment setup for RQ3:} To evaluate the influence of label quality to the quality of model, For MNIST, We modify the specific percentage of labels in the training data randomly (10\% to 100\% at intervals of 10\% and 1\% to 20\% at intervals of 1\%). And for Cifar-10, We modify the specific percentage of labels in the training data randomly (10\% to 60\% at intervals of 10\%).

\textit{Experiment setup for RQ4:} To evaluate the influence of dataset contamination to the quality of model, For MNIST, We slightly modify the contrast of the image to a different direction. And for Cifar-10, We add Gaussian noise, Salt-and-pepper noise to the data set and adjust the brightness of the picture.

\subsection{Experimental Results}

\textit{Results for RQ1:} 

\begin{figure}[]
	\centerline{\includegraphics[width=0.4\textwidth]{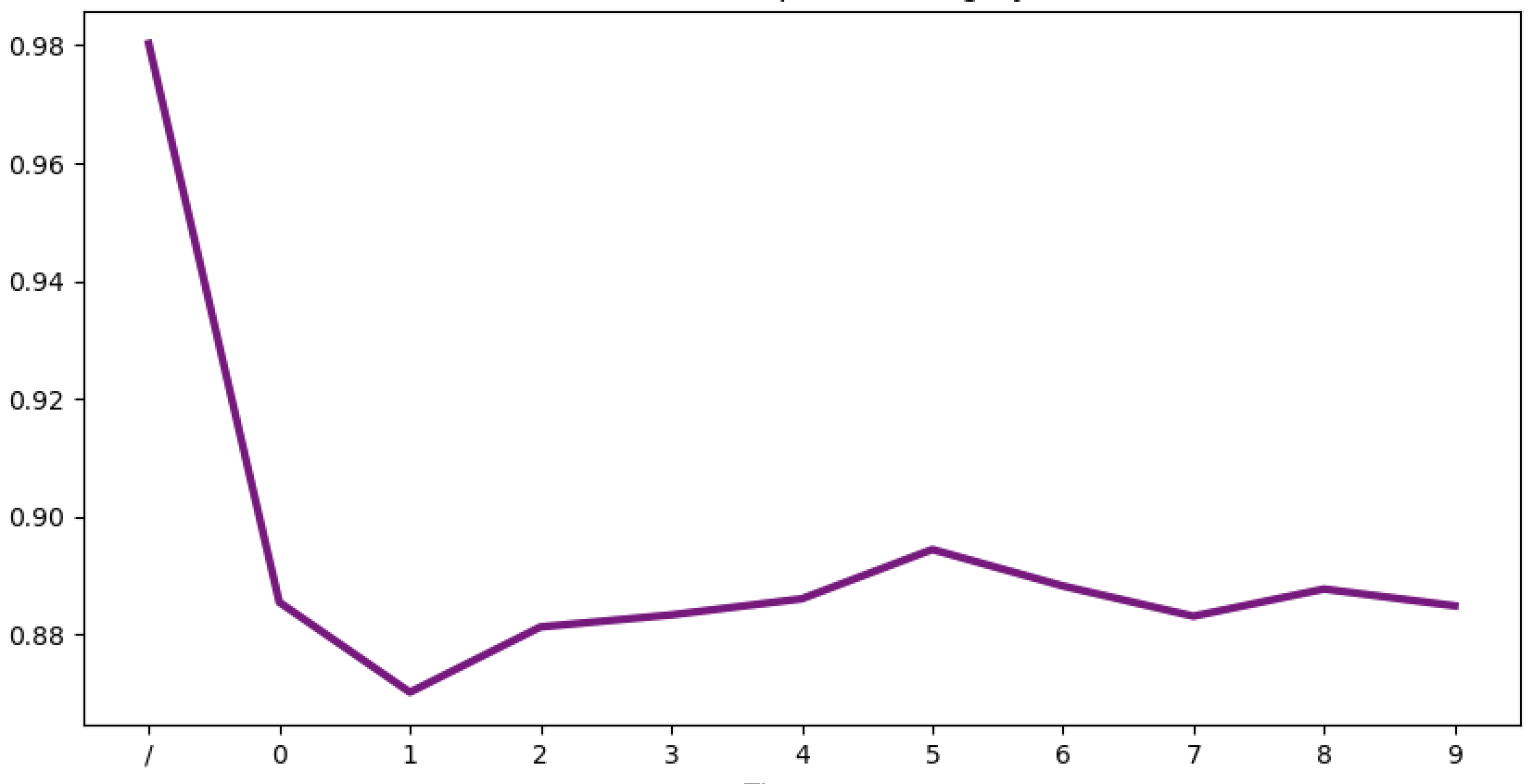}}
	\caption{Accuracy of the model when one specific category is deleted}
	\label{MNIST-balance}
\end{figure}

For MNIST, as can be seen from the result (See Fig. \ref{MNIST-balance}), the testing accuracy decreases significantly when deleting different classes. 

For Cifar-10, we compare the predicted results of the modified training set with those of the completely correct training set. We can find that when we remove pictures of a class or change the labels of the class to that of another, it affects the accuracy of all classes negatively. What's more, different changes cause different results. 
NIN networks show similar results.

\begin{figure}[]
	\centerline{\includegraphics[width=0.4\textwidth]{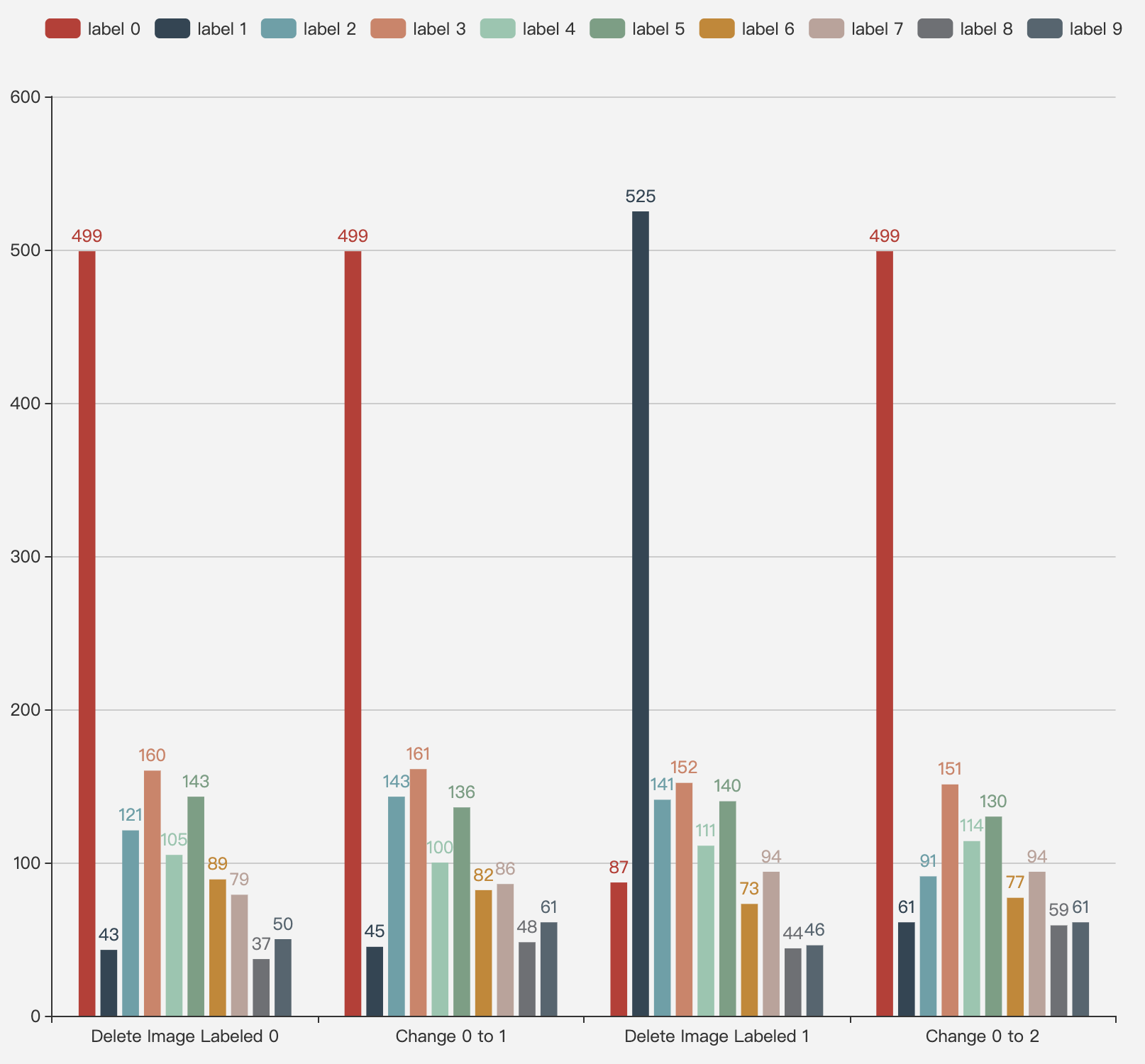}}
	\caption{Labels different from models trained by correct data set(ResNet20)}
	\label{cifar-Eq}
\end{figure}
\begin{figure}[]
	\centerline{\includegraphics[width=0.4\textwidth]{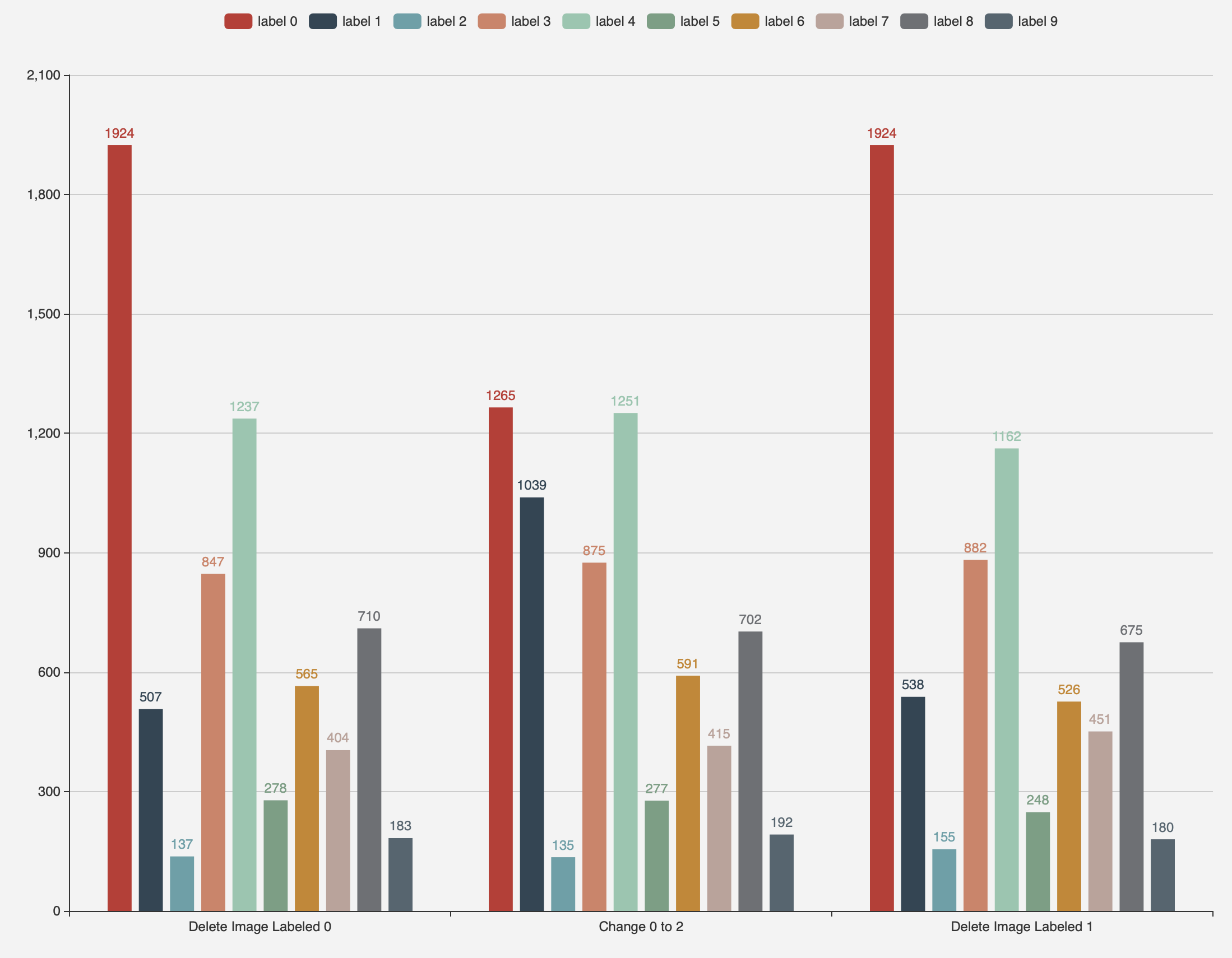}}
	\caption{Labels different from models trained by correct data set(NIN)}
	\label{NIN-p}
\end{figure}

\begin{figure}[]
	\centerline{\includegraphics[width=0.5\textwidth]{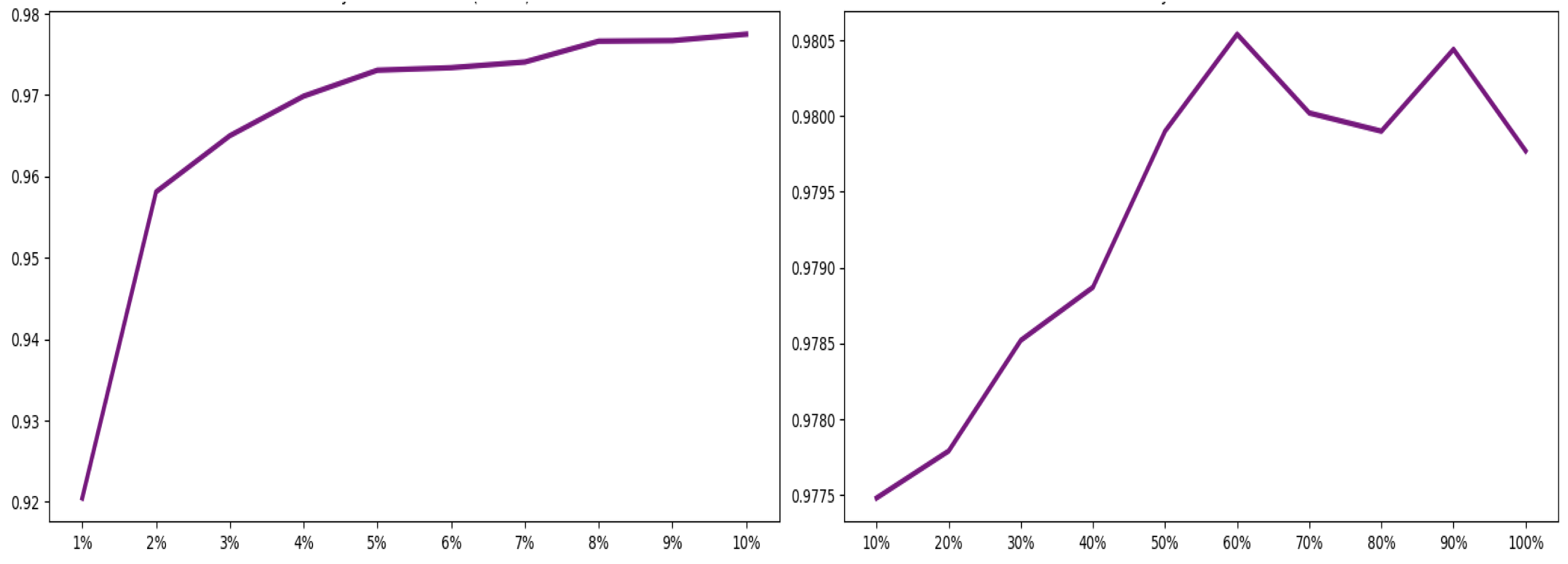}}
	\caption{Accuracy of the model when dataset size are modified (MNIST)}
	\label{MNIST-quantity}
\end{figure}

\textit{Results for RQ2:} 

For MNIST, the size of the training set will not significantly influence the effect of the model and the decrease is not. 

For Cifar-10, we can find that when the scale of the training set is less than 10,000, the accuracy of the testing set trained by the data set drops sharply in Fig. 5. It is safe to say that the larger the data set is, the better the quality of the data set is.What's more, When the test set is large, the number of pictures in training set which used to ensure the quality of the model seems to be certain. When it comes to NIN network: since the model capacity is insufficient, the results of the experiment are not very telling. However, We can also find that when the scale is lower than 10000, it performs bad.

\begin{figure}[]
	\centerline{\includegraphics[width=0.5\textwidth]{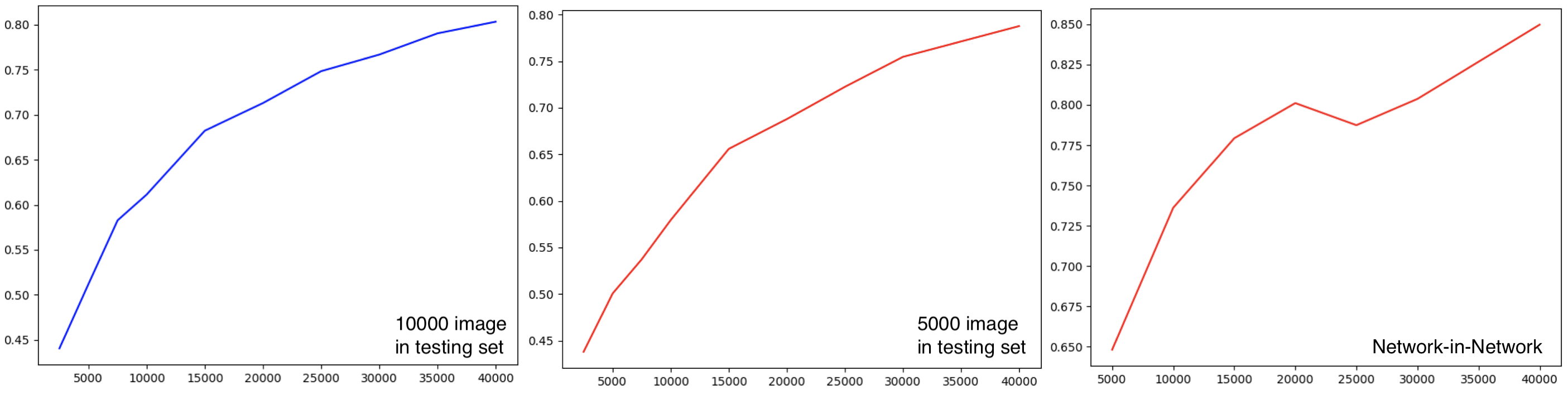}}
	\caption{Modify Dataset Size}
	\label{Cifar-10-size}
\end{figure}

\textit{Results for RQ3:} 

\begin{figure}[htb]
	\centerline{\includegraphics[width=0.45\textwidth]{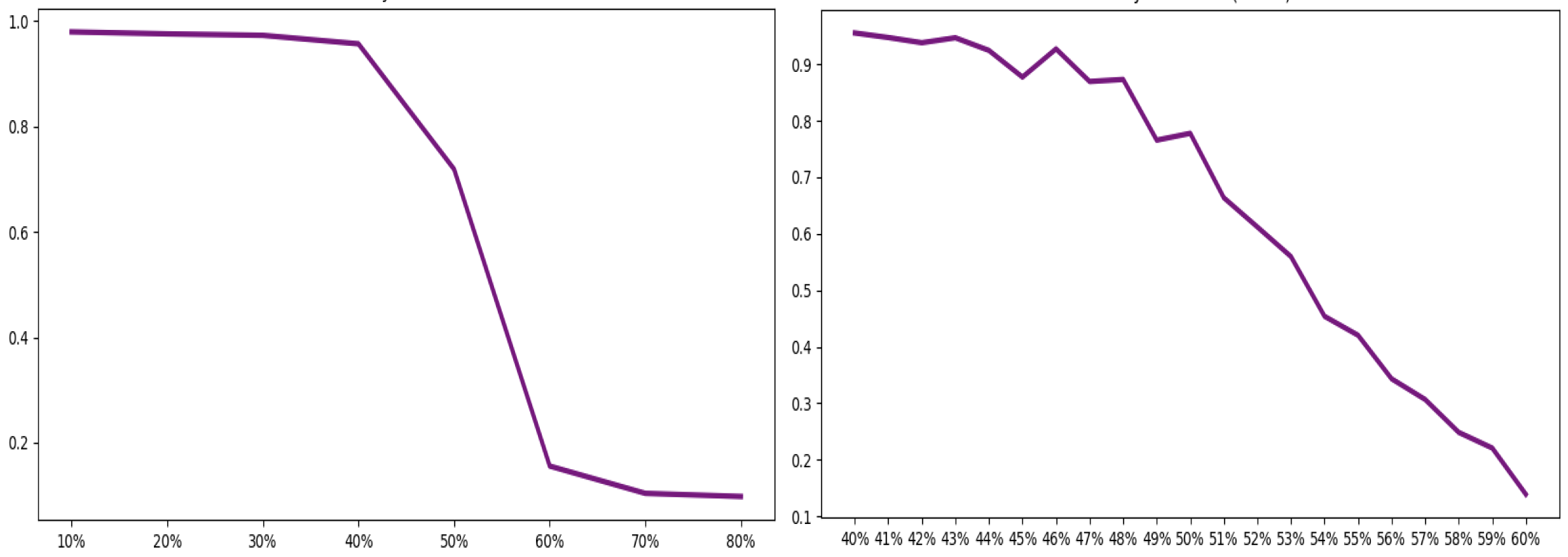}}
	\caption{Accuracy of the model when labels are modified to 0}
	\label{MNIST-label-0}
\end{figure}

\begin{figure}[htb]
	\centerline{\includegraphics[width=0.4\textwidth]{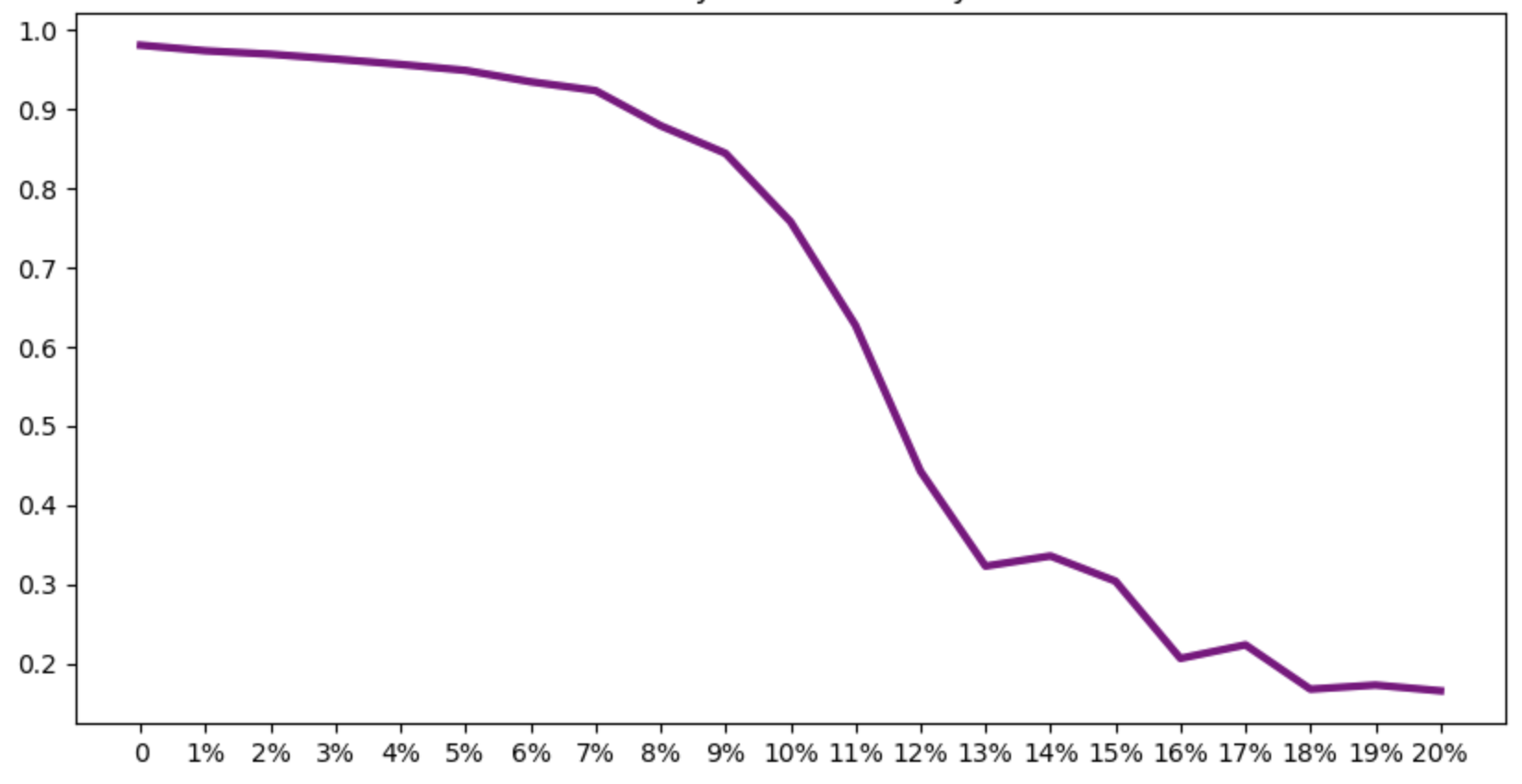}}
	\caption{Accuracy of the model when when labels are modified randomly}
	\label{MNIST-label-r}
\end{figure}

For MNIST, it can be seen from the resulting figure(See Fig. \ref{MNIST-label-0}, Fig. \ref{MNIST-label-r}) that the testing accuracy will decrease to approximately 90\% with the increase of the wrong label percentage. When the wrong label percentage reaches 15\%, the testing accuracy decreased to 10\%, which equals to the excepted value of random judgment. 

\begin{figure}[htb]
	\includegraphics[width=0.5\textwidth]{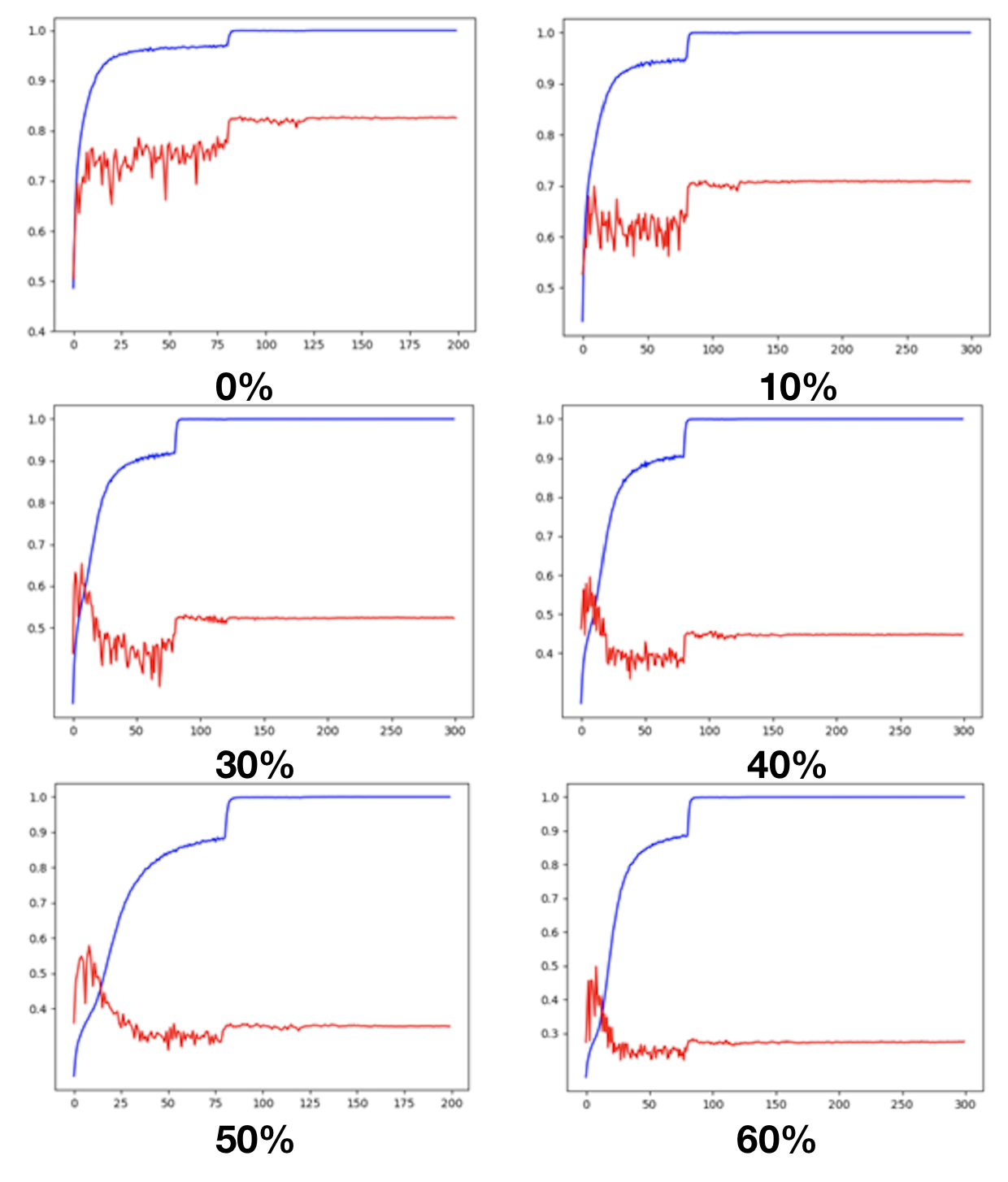}
	\caption{Training set \& testing set accuracy with different label error percentage in Resnet 20}
	\label{Cifar-10-label-acc}
\end{figure}

For Cifar-10, Fig. \ref{Cifar-10-label-acc} shows the effect of label errors on the curve of testing set accuracy. We find that If the error rate of labels is more than 20\%, the accuracy of the testing set obtained declines after a steep increase at the very beginning, and the accuracy rate is not higher than the highest accuracy rate at the beginning of training when accuracy tends to stabilize, which can be considered that more than 20\% error labels are intolerable. When it comes to NIN network, the accuracy also shows that more than 20\% error labels are intolerable. 

\begin{figure}[]
	\centerline{\includegraphics[width=0.3\textwidth]{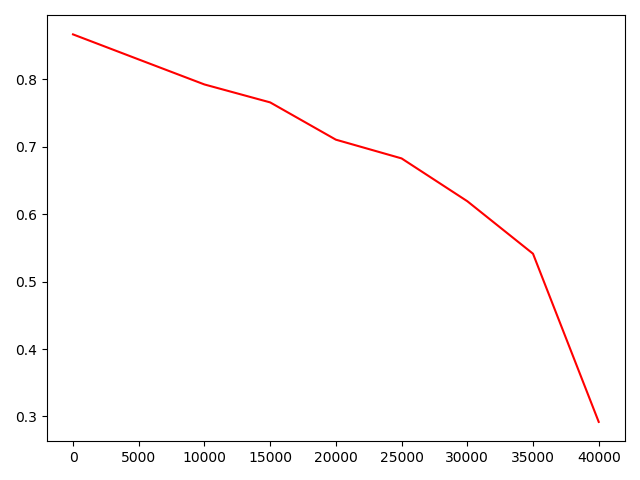}}
	\caption{Accuracy of different label error percentage (NIN)}
	\label{NIN-tea}
\end{figure}

\textit{Results for RQ4:} 

 \begin{figure}[htb]
	\centerline{\includegraphics[width=0.3\textwidth]{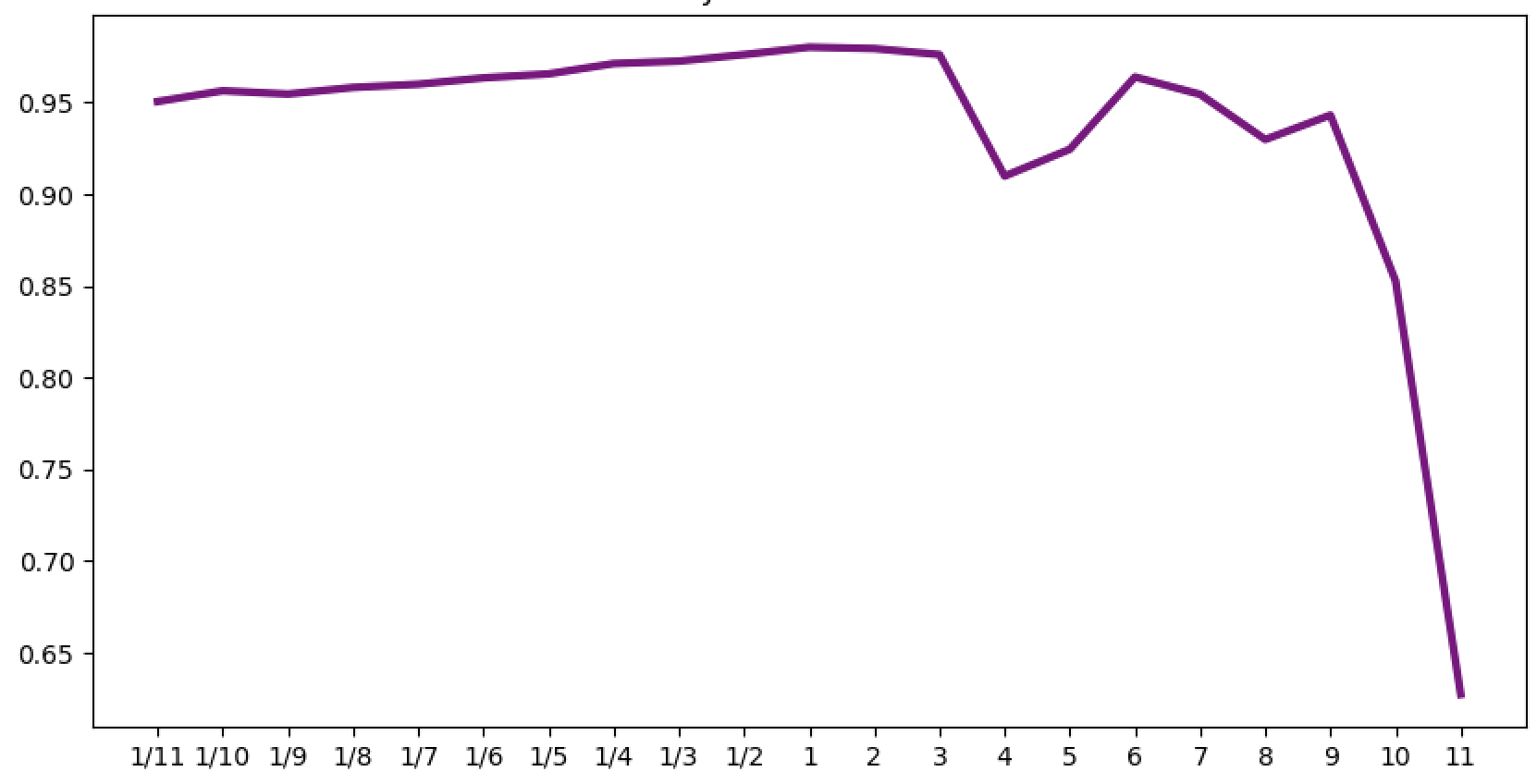}}
	\caption{Accuracy of the model when dataset is contaminated}
	\label{MNIST-contrast-res}
\end{figure}

\begin{figure}[htb]
	\centerline{\includegraphics[width=0.5\textwidth]{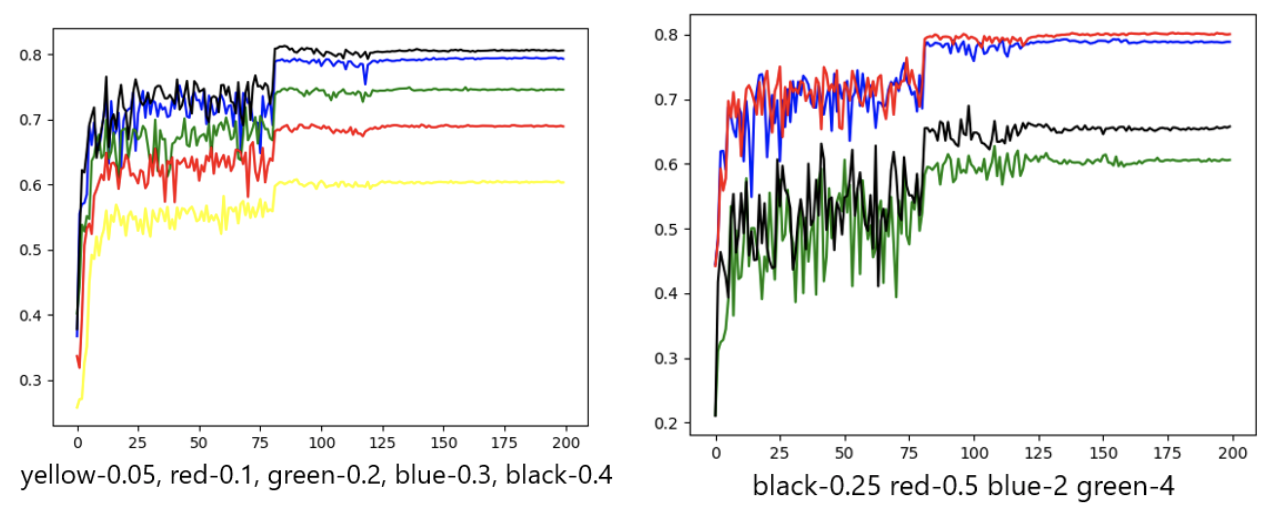}}
	\caption{Accuracy of testing set with Salt-and-pepper \& Brightness noise}
	\label{noise}
\end{figure}

\begin{figure}[htb]
	\centerline{\includegraphics[width=0.5\textwidth]{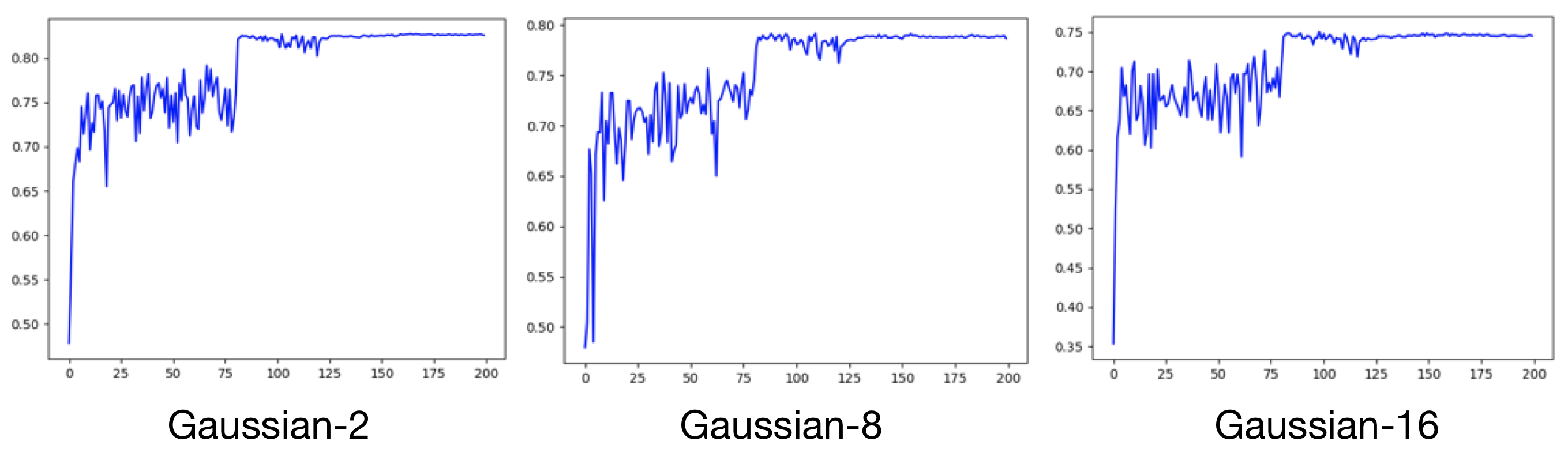}}
	\caption{Accuracy of the dataset with Gaussian noise}
	\label{Gaussian}
\end{figure}

For MNIST, when weakening the contrast, the effect slightly decreases, from 98\% to 95\%. However, when we strengthen the contrast, the effect significantly decreases, from 98\% to 67\%,(See Fig. \ref{MNIST-contrast-res}). One conclusion we get is that the contrast leads to the overfitting problem. 

For Cifar-10, we have the following observations.

\textbf{AWGN:} In our experiment, we use a standard normal distribution to add Gauss noise to the training set. We add the same value of Gauss noise function to the three color paths of each picture as$ f(x) = x + \sigma*random.gauss(0,1)$,  where x is RGB value, and its value is controlled between 0 and 255. $\sigma$ is the multiplier we impose on the normal distribution function. By changing the value of $\sigma$ , We observed the effect of added Gaussian noise on training. We choose 2, 8, 16 as the value of $\sigma$, which is shown in Fig 11. We can find out that when $\sigma$ = 2, the accuracy of the testing set is slightly higher than that of the original dataset which can be considered as the method of avoiding over-fitting is in effect. When it comes to 8 times, things get worse. The quality of the training set has deteriorated when it comes to 16 times. Adding Gaussian noise to dataset does less damage. We also find that for the same $\sigma$, the effect of Gaussian noise on small pictures is greater than that on large ones. Large pictures require high multiples
    
\textbf{Salt-and-pepper noise:}  We add the same value of Salt-and-pepper noise function to the three color paths of each picture as$ f(x) = 0, rand < a $, $ f(x) = x, a < rand < b $, $ f(x) = 255, rand > b $, where x is RGB value and rand is a random value between 0 and 1. We choose 0.05, 0.1, 0.2, 0.3, 0.4 as the value of a and b = 1 - a, We can find in Fig. \ref{noise} that it can't improve the accuracy of the testing set. When the value of a rises to more than 0.2, it may cause a great damage to the dataset. 
	
\textbf{Brightness:} We use the exposure.adjust\textunderscore gamma(x, a) function. x is a matrix of the pictures, and we choose 0.25, 0.5, 2, 4 as the values of a. The bigger the value, the darker the picture is. From the Fig.\ref{noise}, we find that it seems to have an impact on the dataset symmetrically. Darkening the picture is slightly more harmful than brightening it.It is clear that this is not helpful for the increase of testing set accuracy.

\subsection{Remarks and Findings}
\begin{itemize}
    \item \textbf{RQ1:} Dataset equilibrium will affect. Unbalanced data set damages the accuracy of every classification.
    
    \item \textbf{RQ2:} 
    The dataset size will affect. The scale of training set and test set greatly affects the quality of data set. In the current situation, the more pictures the data set has, the better the quality of the dataset is.
    \item \textbf{RQ3:}
    The quality of label will affect.Errors in tags greatly affect the quality of data sets. Adding label errors is considered to be the most effective attack method. What's more, it's easy to attack by changing the label data of the training set.
    \item \textbf{RQ4:} 
    Comparing to other aspects, adding noise to images in training set is less harmful to data set quality. Gaussian noise and Salt-and-pepper noise can help raise the accuracy of the testing set, while the change of brightness seems helpless.
    Adding noise properly can help to reduce generalization error and improve the accuracy of test set.
\end{itemize}

\section{Conclusion and Future Work}
In this work, we study the four aspects that affect the quality of datasets in detail and give the experimental results for Cifar-10 and MNIST. We believe that most of the conclusions of the experiment are universal. Some modifications to the image itself may vary from dataset to dataset. In the next study, we will consider the four aspects separately and carefully and determine a universal distance function to measure the quality of datasets.

\bibliography{main}

\end{document}